%% file: main.tex
\documentclass[letterpaper, 10 pt, conference]{ieeeconf}  

\IEEEoverridecommandlockouts                              

\input{preamble.tex}

\input{commands.tex}

\usepackage{bm}
\newcommand{\citet}[1]{\cite{#1}}
\newcommand{\citep}[1]{\cite{#1}}

\title{\LARGE \bf \titlelong{}}

\author{Artemii Redkin$^{1}$, Zdravko Dugonjic$^{1}$, Mike Lambeta$^{2}$, Roberto Calandra$^{1}$
\thanks{$^{1}$LASR Lab, TU Dresden, Dresden, Germany}
\thanks{$^{2}$Meta AI, Menlo Park, CA, USA}
}

\begin{document}

\maketitle
\thispagestyle{empty}
\pagestyle{empty}


\begin{abstract}
	\input{0_abstract.tex}
\end{abstract}



\section{INTRODUCTION}
	
	\input{1_introduction.tex}


\section{RELATED WORK}
\label{sec:related}

	\input{2_related.tex}


\section{BACKGROUND}
\label{sec:background}

	\input{2_background.tex}


\section{DYNAMIC ILLUMINATION FOR VISION-BASED TACTILE SENSORS} 
\label{sec:approach}

\input{3_approach.tex}


\section{EXPERIMENTAL RESULTS}
\label{sec:result}

	\input{4_result.tex}


\section{CONCLUSION}
\label{sec:conclusion}

\input{5_conclusion.tex}







\section*{ACKNOWLEDGMENT}
\input{99_acknowledgments.tex}


\bibliographystyle{IEEEtran}
\bibliography{references}

\end{document}

%% file: preamble.tex
\usepackage[english]{babel}

\usepackage{graphicx}
\usepackage{animate}
\usepackage{epsfig} 				
\usepackage{epstopdf}

\usepackage{listings}
\usepackage{color}
\usepackage{nameref}
\usepackage{hyperref}
\usepackage{amsmath}	 			
\usepackage{amssymb}  				

\usepackage{dsfont}			
\usepackage{mathtools}

\usepackage{epigraph}
\usepackage{lscape}
\usepackage[]{nomencl}				
\usepackage{algorithm}
\usepackage{algorithmic}
\usepackage{multicol}
\usepackage{multirow}
\usepackage{etoolbox}

\usepackage{caption}
\usepackage{subcaption}
\usepackage{wrapfig}

\usepackage{siunitx}

\usepackage{floatflt}

\usepackage{url}

%% file: commands.tex
\newcommand{\email}[1]{\href{mailto:#1}{\nolinkurl{#1}}}

\newcommand{\fig}[1]{Fig.~\ref{#1}}

\newcommand{\titlelong}[0]{Enhance Vision-based Tactile Sensors\\ via Dynamic Illumination and Image Fusion}

%% file: 0_abstract.tex
Vision-based tactile sensors use structured light to measure deformation in their elastomeric interface.
Until now, vision-based tactile sensors such as DIGIT and GelSight have been using a single, static pattern of structured light tuned to the specific form factor of the sensor.
In this work, we investigate the effectiveness of dynamic illumination patterns, in conjunction with image fusion techniques, to improve the quality of sensing of vision-based tactile sensors.
Specifically, we propose to capture multiple measurements, each with a different illumination pattern, and then fuse them together to obtain a single, higher-quality measurement. 
Experimental results demonstrate that this type of dynamic illumination yields significant improvements in image contrast, sharpness, and background difference.
This discovery opens the possibility of retroactively improving the sensing quality of existing vision-based tactile sensors with a simple software update, and for new hardware designs capable of fully exploiting dynamic illumination.

%% file: 1_introduction.tex
 In robotics, haptic exploration is central to understanding the world through touch interactions~\citep{Lederman1987Hand}. 
Tactile sensors allow robots to collect essential information about their surroundings, precisely manipulate objects, and ensure safe interactions within dynamic environments~\citep{calandra2017feeling}. 
By detecting physical contact, tactile sensing allows robots to avoid collisions, adjust movements, and handle objects delicately, especially in tasks that require fine interactions~\citep{lambeta2020digit,qi2023general}.
 
Vision-based Tactile Sensors (VBTS)  are a popular choice of tactile sensors~\citep{yuan2014tactile,WardCherrier2018TacTip,lambeta2020digit}. 
They enable robots to perceive their environment by capturing surface deformations upon contact with objects, thus facilitating the measurement of forces, textures, and shapes. 
VBTS typically incorporate structured light in their construction, and currently, all such sensors use static illumination, meaning the lighting intensity and colors remain constant during measurements.

Enhancing images from VBTS holds pivotal importance due to their widespread applicability across diverse robotic tasks. 
These sensors serve as crucial components in robotic systems, providing essential data for various operations. 
The state-of-the-art approach involves training deep neural networks using images from VBTS, where the quality of the input image significantly influences the model's performance and output. 
Improved imaging quality from VBTS could offer deeper insights into robotic interactions with objects, ultimately enhancing problem-solving capabilities. 
Addressing this need, our study aims to explore the feasibility of image enhancement in VBTS and propose methodologies for achieving this enhancement.

\begin{figure}[t]
    \centering
    \includegraphics[width=0.98\linewidth]{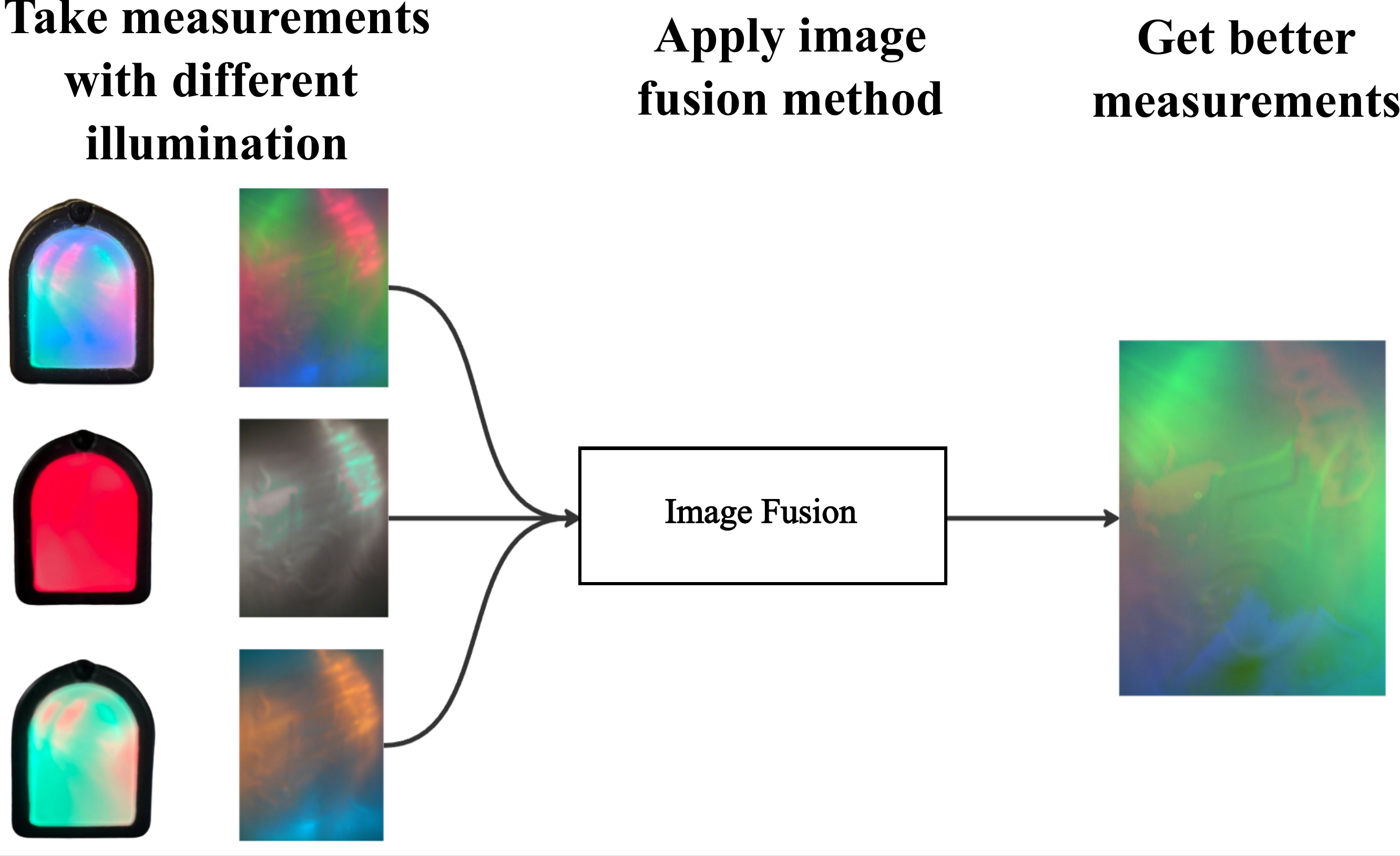}
    \caption{Current vision-based tactile sensors use static illumination patterns. In this work, we instead propose to collect several measurements under dynamic illumination conditions, and then fuse them together in a single higher-quality measurement. Experimental results show that this approach yields significantly improved quality of sensing.}
    \label{fig:teaser}
\end{figure}

In this study, we contribute to the field by establishing a framework to enhance the measurement quality of vision-based tactile sensors through the application of dynamic lighting and image fusion techniques (\fig{fig:teaser}). 
Our investigation delves into the mathematical formulation of this framework, and the comprehensive evaluation and demonstration of diverse approaches tailored to enhance image quality. 
Specifically, our methodology integrates dynamic lighting schemes to enhance contrast and sharpness, while employing image fusion algorithms to combine multiple sensor outputs into cohesive images.
We further validate the feasibility of enhancing sensor images and conduct a comparative analysis of various illumination variations and image fusion methods, assessing their applicability to vision-based tactile sensors. 
Through rigorous experimentation and analysis, we present a spectrum of effective techniques poised to enhance images acquired from VBTS.

The development of techniques for enhancing images from VBTS holds promise in advancing the capabilities of robotic systems. 
By improving image quality, this research equips robots with deeper insights into their interactions with objects, thereby enhancing their problem-solving abilities across a set of tasks. 
Our systematic exploration and validation of these enhancement techniques lay a solid foundation for the integration of advanced imaging capabilities into robotic systems. 
This paves the way for more efficient and effective robotic applications in various real-world scenarios, thereby contributing significantly to the advancement of robotics technology.

Our contributions are:
\begin{itemize}
    \item We introduce an approach of dynamic lighting for vision-based tactile sensors and demonstrate the methodology for its usage.
    \item Show that it is possible to enhance the measurements from the sensor using dynamic lighting and image fusion techniques.
    \item Identify the most effective image fusion method to be used in conjunction with dynamic lighting.
    \item Determine the number of images for optimal output image quality.
    \item Analyze the time required to apply dynamic lighting effectively.
\end{itemize}


%% file: 2_related.tex
\subsection{Illumination in Vision-Based Tactile Sensors}
Previous research in the field of vision-based tactile sensing focused on the strategic positioning of lighting systems \textit{at design time}, ensuring that illuminated elastomer gives the optimal response for downstream tasks. 
\cite{do2023densetact} noted that more light sources improve tactile readings, allowing better light distribution over the elastomer surface. 
In \cite{azulay2023allsight}, it is evaluated how three different illumination setups affect the performance of the contact state estimation problem. 
\cite{wang2021gelsight}, motivated by the design of their new sensor, compares how both positioning and combinations of monochrome red, green, and blue lights impact the results of the 3D reconstruction task. 
\cite{sun2022soft} showed that removing the color from the structured lights negatively affects the force prediction.
\cite{Lambeta2024Digitizing} introduced a simulation approach to perform a careful study of design parameters of the optical and illumination system for an omnidirectional VBTS.
Unlike prior research, our study systematically evaluates the effect of combining images captured under dynamic illumination setups compared to static ones.

More similar to our work is \cite{johnson2011microgeometry} which sequentially turned on a single light out of the 6 placed at the circumference of the sensor.
Subsequently, the black and white images were used to reconstruct the surface of the object using the shadows in a photometric stereo setting.
Compared to this work, our approach relies on machine learning tools to process the images, thus being less sensitive to strong assumptions such as the known illumination model, and the linearity of the model.

\subsection{Active Illumination for Photogrammetry}
While the aforementioned work focuses on static lighting configurations, a broader field in computer vision demonstrates the advantages of active lighting. 
Based on the idea of photometric sampling \cite{nayar1989determining}, the authors in \cite{wenger2005performance} proposed the method for obtaining object reflectance and surface normal. 
They achieved that by recording the scene illuminated with high-frequency pulsing LED light sources placed around the object. 
\cite{raskar2004non} shows that it is possible to create a depth edge map by flashing the scene with lights around the camera lens.
More recently, one notable example of dynamic lighting is the quantitative differential phase contrast imaging technique introduced in~\cite{tian2015quantitative}. 
This method utilizes different lighting conditions in an LED array microscope to enhance phase contrast, improving the visualization of transparent samples in biological research without requiring complex optical setups.
Unlike traditional applications focused on visual imaging, microscopy, or medical diagnostics, applying these techniques to tactile sensing introduces innovative strategies for capturing and interpreting tactile information.

\subsection{Image Fusion}

In the domain of image fusion, \cite{wang2011laplacian} proposed the Laplacian Pyramid method, addressing multi-focus image fusion by decomposing images into multiple levels and selectively incorporating focused elements from each level into the final image.
This technique preserves the best-focused aspects of each original image, which is particularly beneficial in fields where detailed texture information is essential.
Further advancements in image fusion include the discrete fractional wavelet transform method introduced in~\cite{xu2016medical}. 
This approach allows for integrating multiple medical images into a single composite, retaining critical information from each source image for improved medical diagnosis and treatment planning.
However, previous research has not explored enhancing the quality of images in the context of vision-based tactile sensors.

%% file: 2_background.tex
\subsection{Vision-based Tactile Sensors}

Although many VBTS have been introduced in the literature \citep{yuan2014tactile,WardCherrier2018TacTip,lambeta2020digit}, here we focus on explaining the working principle of the widespread DIGIT sensor \cite{lambeta2020digit} which we use in our experiments.
DIGIT is compact and versatile design which allows easy integration into various robotic platforms, while its durability and cost-effectiveness ensure long-term value. 
These features, coupled with the sensor's ability to handle delicate tasks and navigate complex environments, establish DIGIT as a popular choice for advanced robotic applications, offering a balance of performance, adaptability, and affordability. 
The DIGIT sensor comprises the following components:

\begin{enumerate}
\item Elastomer Skin: A deformable surface.
\item Embedded Camera: Strategically positioned to capture images of the elastomer's deformations upon contact.
\item Illumination System: Ensures consistent lighting conditions for clear image capture.
\item Compact Housing: All components are encased in lightweight housing, facilitating seamless integration with robotic systems.

\end{enumerate}

The core mechanism of vision-based tactile sensors revolves around detecting changes on the sensor's contact surface. 
The embedded camera captures the elastomer deformations when it interacts with an object. 
By analyzing these images, it is possible to deduce the forces applied, the shape of the object in contact, and other properties like texture. 
This information proves valuable across a spectrum of robotic applications, including object recognition, grip control, and manipulation.

\subsection{Image Fusion}
Image fusion is the process of combining two or more images into a single composite image that integrates and preserves the most important information from each of the individual images. 
Image fusion can be defined as a mapping function \( f: \{ I_1, I_2, \dots, I_n \} \rightarrow I^*,\) where \( I_1, I_2, \dots, I_n \) is a set of input images, \( I^* \) -- the fused image that contains integrated information from all input images, function \( f \) -- the image fusion algorithm, designed to preserve or enhance relevant features from the input images.
We now discuss several image fusions techniques that are evaluated in our experiments:

\subsubsection{Channel-wise Summation}
Channel-wise Summation is defined as $I^* = RGB(I_R, I_G, I_B)$ where $RGB(I_R, I_G, I_B)$ represents combining the red channel from $I_R$, the green channel from $I_G$, and the blue channel from $I_B$ to form the resulting image.

\subsubsection{Brovey Fusion}
The modification of the Brovey Fusion we used is defined as $I^*=n_R*I_R+n_G*I_G+n_B*I_B$ where $I_R$, $I_G$, and $I_B$ represent the red, green, and blue channels of the input image, respectively, $n_k$ represents the normalized value of the pixel of image $I_k$.

\subsubsection{Laplacian Pyramid}
The Gaussian pyramid \cite{adelson1980image} is a multi-scale representation of an image, which is constructed by applying a series of Gaussian filters and downsampling the image iteratively. 
Creating a Gaussian pyramid of an image involves a series of steps where each level of the pyramid is a lower resolution version of the previous level.
Given an input image \(I\), it undergoes convolution with a Gaussian kernel 
\(G\), defined as
\begin{equation*}
    G(x, y) = \frac{1}{2\pi\sigma^2}e^{-\frac{x^2 + y^2}{2\sigma^2}}\,,
\end{equation*}
where \(\sigma\) is the standard deviation of the Gaussian distribution, and \(x\) and \(y\) are the distances from the center of the kernel. 
This convolution operation can be represented as $I' = I * G$, where \(I'\) is the smoothed image, and \( * \) denotes the convolution operation. The smoothed image is then subsampled, reducing its resolution by typically retaining every second pixel in both horizontal and vertical directions $I''(x, y) = I'(2x, 2y)$ where $I' = I * G$. 
This reduces the number of pixels in the image by a factor of 4, halving both the image's width and height. 
This process of smoothing and subsampling is iteratively repeated for multiple levels, yielding the hierarchical structure known as the Gaussian pyramid $I_n = (I_{n-1} * G)_{\downarrow 2}$, where \(I_0\) is the original image, \(\downarrow 2\) denotes subsampling (taking every second pixel), and \(I_{n-1}\) is the image at the previous level of the pyramid.

The Gaussian pyramid is used as a foundational step in creating the Laplacian pyramid.
Given an image \(I\), a Laplacian pyramid~\cite{burt1987laplacian} is constructed to encode the image at multiple levels of resolution, focusing on the image details. 
To create a Laplacian pyramid of the image \(I\), first, the Gaussian pyramid  is constructed, denoted as \(G_0, G_1, \ldots, G_n\), where \(G_0\) is the original image, and \(G_i\) is the \(i\)-th level Gaussian pyramid image.
The Laplacian pyramid levels \(L_0, L_1, \ldots, L_{n-1}\) are calculated as $L_i = G_i - \text{Expand}(G_{i+1})$ for each level \(i\), where \(\text{Expand}(G_{i+1})\) upsamples \(G_{i+1}\) and then convolves it with the Gaussian kernel. 
This process reveals the details that differ between \(G_i\) and the approximation of \(G_i\) from \(G_{i+1}\).
The last level of the Laplacian pyramid, \(L_n\), is simply $L_n = G_n$. 
To perform image fusion with Laplacian pyramid for images \(I_1, I_2, \ldots, I_n\) and obtain a composite image \(I^*\), the Laplacian pyramids \(\{L^i_1, L^i_2, \ldots, L^i_k\}\) of all the input images need to be constructed.
Then create a composite Laplacian pyramid \(\{C_1, C_2, \ldots, C_k\}\) by fusing corresponding levels
$C_j(x, y) = F(L^1_j(x, y), L^2_j(x, y), \ldots, L^n_j(x, y))$ and reconstruct the composite image \(I^*\) from the composite Laplacian pyramid
$I^* = C_k + \sum_{j=1}^{k-1}{Upscale}(C_j)$, where the up-scaling operation, denoted as \(\text{Upscale}(\cdot)\), increases an image's resolution from the composite Laplacian pyramid prior to combining it with the next higher level. This process involves interpolating the image to a resolution that aligns with the next pyramid level and applying a low-pass filter to reduce high-frequency artifacts introduced by interpolation.

\subsubsection{Discrete Wavelet Transform (DWT) Fusion}
The wavelet transform \cite{kingsbury1998wavelet} is a mathematical tool used in signal processing \cite{daubechies1990wavelet} and image analysis for decomposing a signal or an image into its constituent parts at different scales. 
The wavelet transform provides a multi-resolution analysis by representing the image in terms of a set of basis functions, called wavelets, which are localized in both space and frequency.

Given an image \(I(x, y)\), the two-dimensional discrete wavelet transform (DWT) decomposes the image into its constituent parts at different scales. 
As a first step of the DWT decomposition process, a low-pass filter is applied \(L\) and a high-pass filter \(H\) to each row of the image \(I\), followed by down-sampling by 2:
\begin{equation*}
\begin{aligned}
    I_{L}^{horizontal}(x, y) &= \text{Downsample}\left(I(x, y) * L\right) \\
    I_{H}^{horizontal}(x, y) &= \text{Downsample}\left(I(x, y) * H\right).
\end{aligned}\,
\end{equation*}
Then, apply the same pair of filters to the columns of the horizontally filtered images, followed by downsampling by
\begin{equation*}
\begin{aligned}
LL(x, y) &= \text{Downsample}\left(I_{L}^\text{horizontal}(x, y) * L\right)\,,\\
LH(x, y) &= \text{Downsample}\left(I_{L}^\text{horizontal}(x, y) * H\right)\,,\\
HL(x, y) &= \text{Downsample}\left(I_{H}^\text{horizontal}(x, y) * L\right)\,, \\
HH(x, y) &= \text{Downsample}\left(I_{H}^\text{horizontal}(x, y) * H\right)\,.
\end{aligned}\,
\end{equation*}

This results in four sub-bands: \(LL\) (approximation), \(LH\) (horizontal detail), \(HL\) (vertical detail), and \(HH\) (diagonal detail). The \(LL\) sub-band can be further decomposed using the same process to achieve more levels of detail and approximation. 
Discrete wavelet image fusion involves applying wavelet transforms to input images, decomposing them into approximation and detail coefficients. These coefficients are fused using selected rules such as maximum, minimum, or weighted average. 
Fused coefficients are then used to reconstruct a composite image via inverse wavelet transform. 
By selectively combining coefficients, the resulting image retains essential features from the input images while minimizing artifacts, offering a comprehensive representation of the original data.

%% file: 3_approach.tex
\subsection{Task definition}
The objective of image fusion is to combine two or more images into a single output that enhances overall image quality, making the selection of an effective fusion method essential. 
With dynamic lighting approach, image fusion is closely linked with determining the optimal illumination patterns for the touch sensor. 
Both identification of optimal illumination patterns of the tactile sensor,  number of images required and selection of the most effective image fusion fusion method are crucial in dynamic lighting.
This problem can be considered as the optimization task
\begin{equation}
\underset{\Theta, n, f} {argmax}\ \mathcal{P}\left(f\left(I_{\theta_1}, I_{\theta_2}, \ldots, I_{\theta_n}\right)\right)\,,
\end{equation}
where $\Theta = \{\theta_1, \theta_2, \ldots, \theta_n\} $ represents the set of illumination patterns in which the images $ I_{\theta_i} $ were taken, $n$ is the image budget - the number of images to be used for fusion, $ f: (I_1, I_2, \ldots, I_n) \mapsto I^* $ is the image fusion method, $\mathcal{P}: I^* \mapsto \mathbb{R} $ is some image quality metric applied to the resulting fused image, and $I_{\Theta_i} $ is the  image  taken in illumination determined by $\theta_i $.

\subsection{Metrics}
An important question regarding the formulation above is: \textit{what image quality metric should we use?}
Unfortunately, this question does not have a satisfyng answer since it might depend from the downstream task we might care about.
Without loss of generalization of the problem formulation, in our experiments, we use several common metrics to evaluate image quality:
\subsubsection{Gradient-based Sharpness}
Gradient-based Sharpness is defined as $S = \frac{1}{N} \sum_{i=1}^{N} \sqrt{\left(\frac{\partial I}{\partial x}\right)^2 + \left(\frac{\partial I}{\partial y}\right)^2}$,where $I$ represents the image, $N$ - the number of pixels of the image, $\frac{\partial I}{\partial x_i}$ and $\frac{\partial I}{\partial y_i}$ are the partial derivatives of the image intensity with respect to the spatial coordinates $x_i$ and $y_i$.

\subsubsection{Root Mean Squared Contrast}
Root Mean Squared Contrast is defined as $C_{rms} = \sqrt{\frac{1}{N} \sum_{i=1}^{N} (I_i - \mu)^2},$
where $I_i$ represents the intensity of the $i$-th pixel, $\mu$ is the mean intensity of all pixels, $N$ is the total number of pixels in the image.

\subsubsection{Difference with Background} Difference with Background is defined as $D = \frac{1}{N} \sum_{i=1}^{N} |I_i - B_i|$, where $I$ is the image of the elastomer's surface in contact with an object and $B$ is the background image (i.e., the image obtained from the sensor without touching any object).

%% file: 4_result.tex
\begin{figure}[t]
  \centering
  \includegraphics[width=\linewidth]{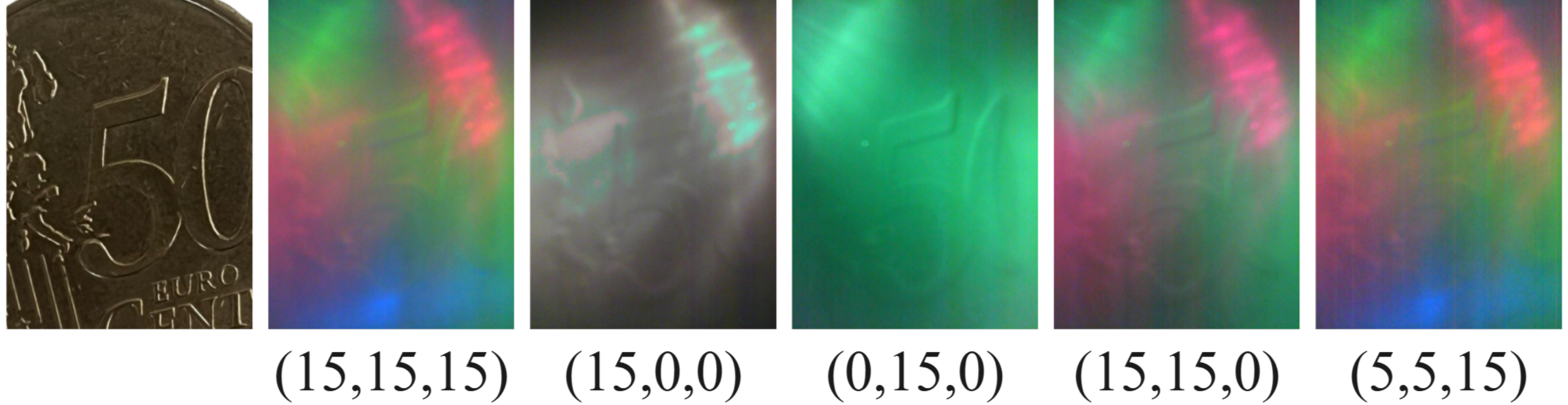}
  \caption{Images of a coin, and corresponding measurements obtained with DIGIT with different illumination settings.}
\label{fig:coin_different_illumination}
\end{figure}

\begin{figure}[t]
  \centering
  \includegraphics[width=\linewidth]{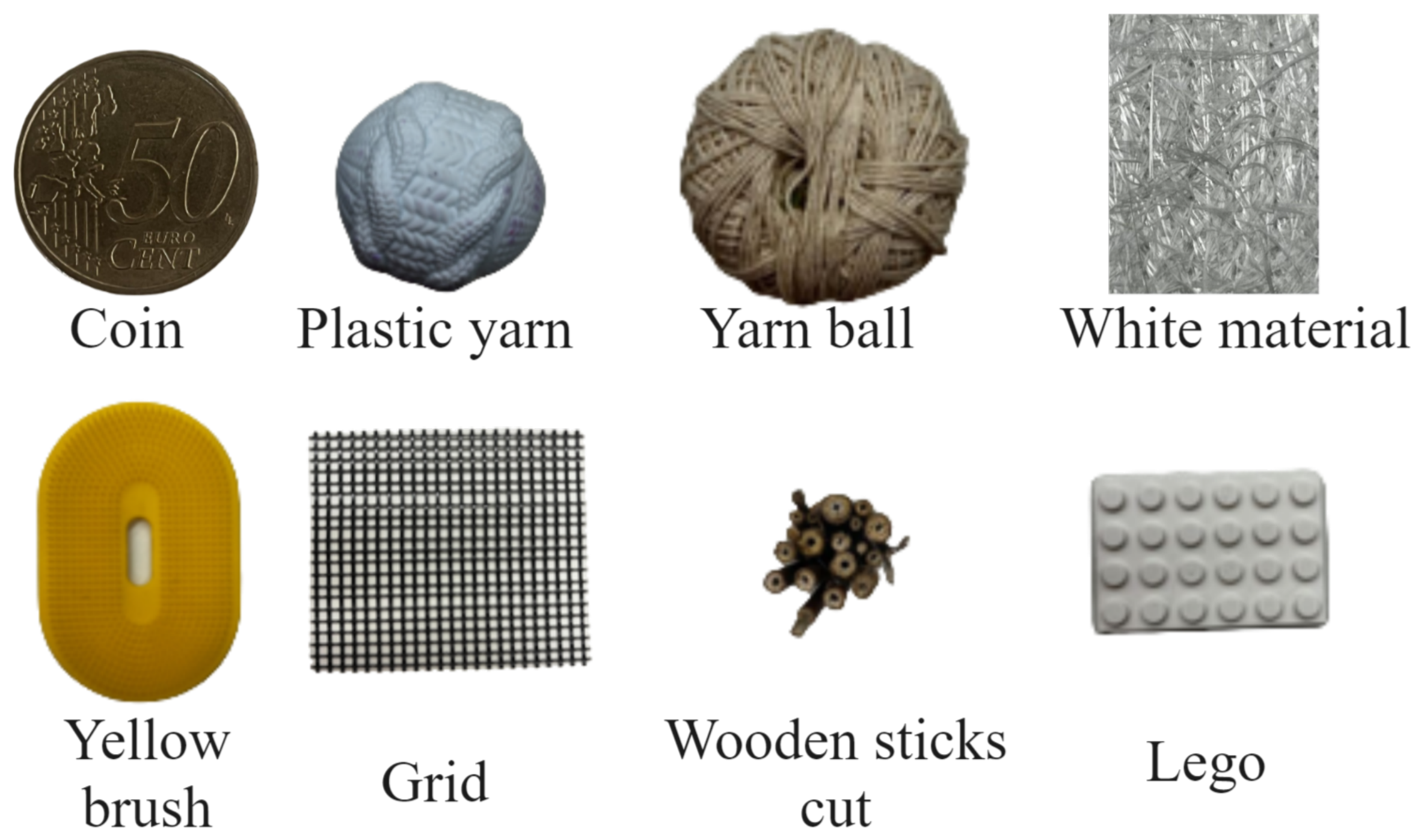}
  \caption{Objects used in the experiments.}
  \label{fig:objects}
\end{figure}

In the experimental evaluation, we aim to answer the following questions:
\begin{itemize}[leftmargin=*]
    \item Can we enhance the quality of measurements for a DIGIT sensor using dynamic lighting and image fusion techniques? 
    \item Can we improve all selected metrics simultaneously? 
    \item What is the most effective fusion method for dynamic lighting? 
    \item What is the temporal cost of dynamic lighting?
\end{itemize}
For our experiments, we employed a standard DIGIT~\citep{lambeta2020digit} vision-based tactile sensor equipped with three LED lights: red, green, and blue. 
The intensity of each light can be adjusted from 0 to 15, where 0 means no light and 15 - maximum intensity, enabling the creation of various illumination patterns represented by tuples (R, G, B), where R, G, and B denote the intensity values of the red, green, and blue LEDs, respectively. 
By default, the DIGIT sensor is set to (15, 15, 15), with all LEDs operating at maximum intensity.
To facilitate experiments, we mounted the sensor on a fixed frame, enabling us to capture multiple images from the sensor while maintaining a consistent spatial position w.r.t. the touched objects.
Throughout our experiments, we used eight different objects with different tactile properties, as shown in \fig{fig:objects}.

\begin{figure*}[t]
  \centering
    \begin{subfigure}{0.49\linewidth}
        \centering
        \includegraphics[height=5cm]{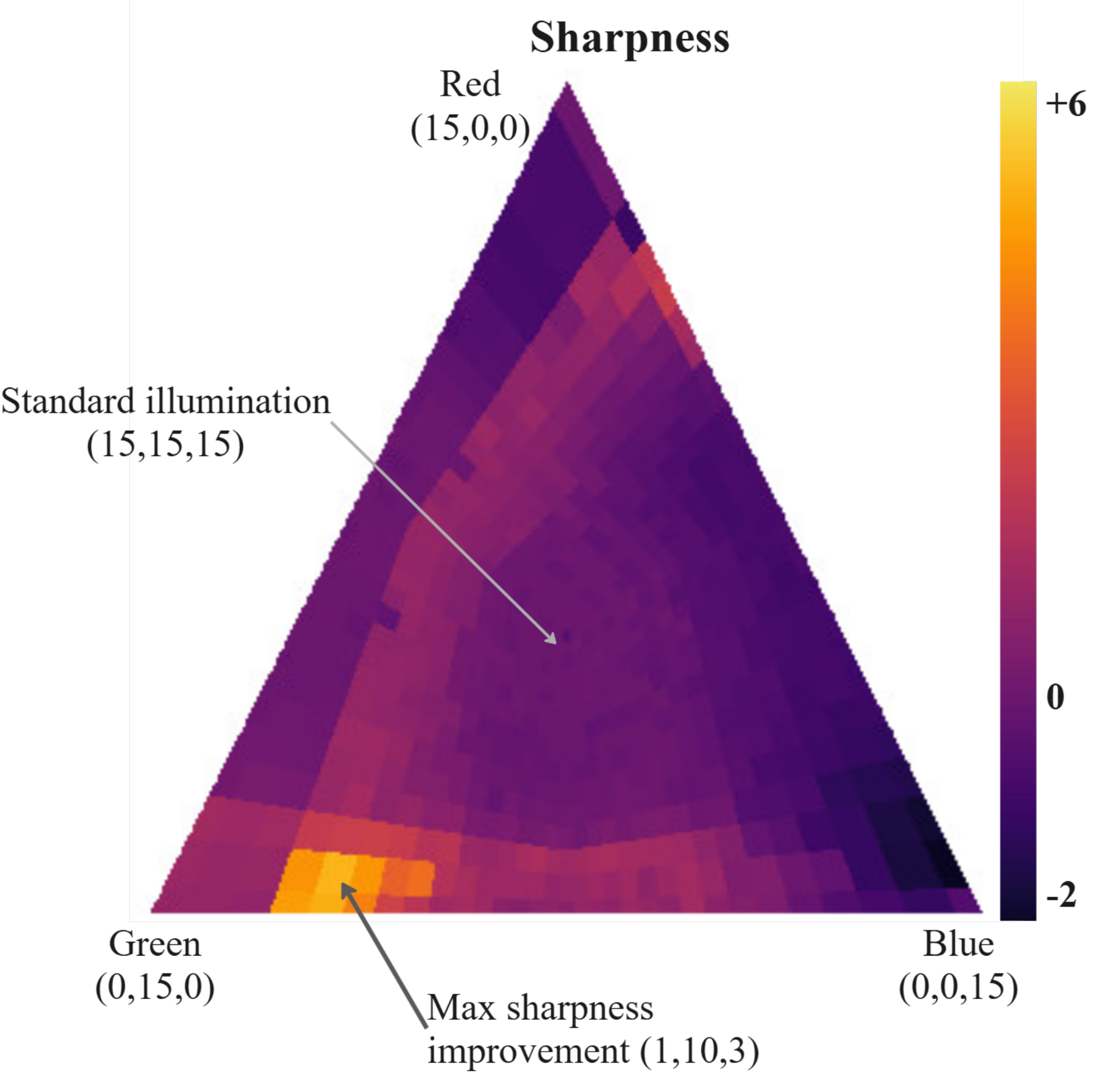}
    \end{subfigure}
    \hfill
    \begin{subfigure}{0.49\linewidth}
        \centering
        \includegraphics[height=5cm]{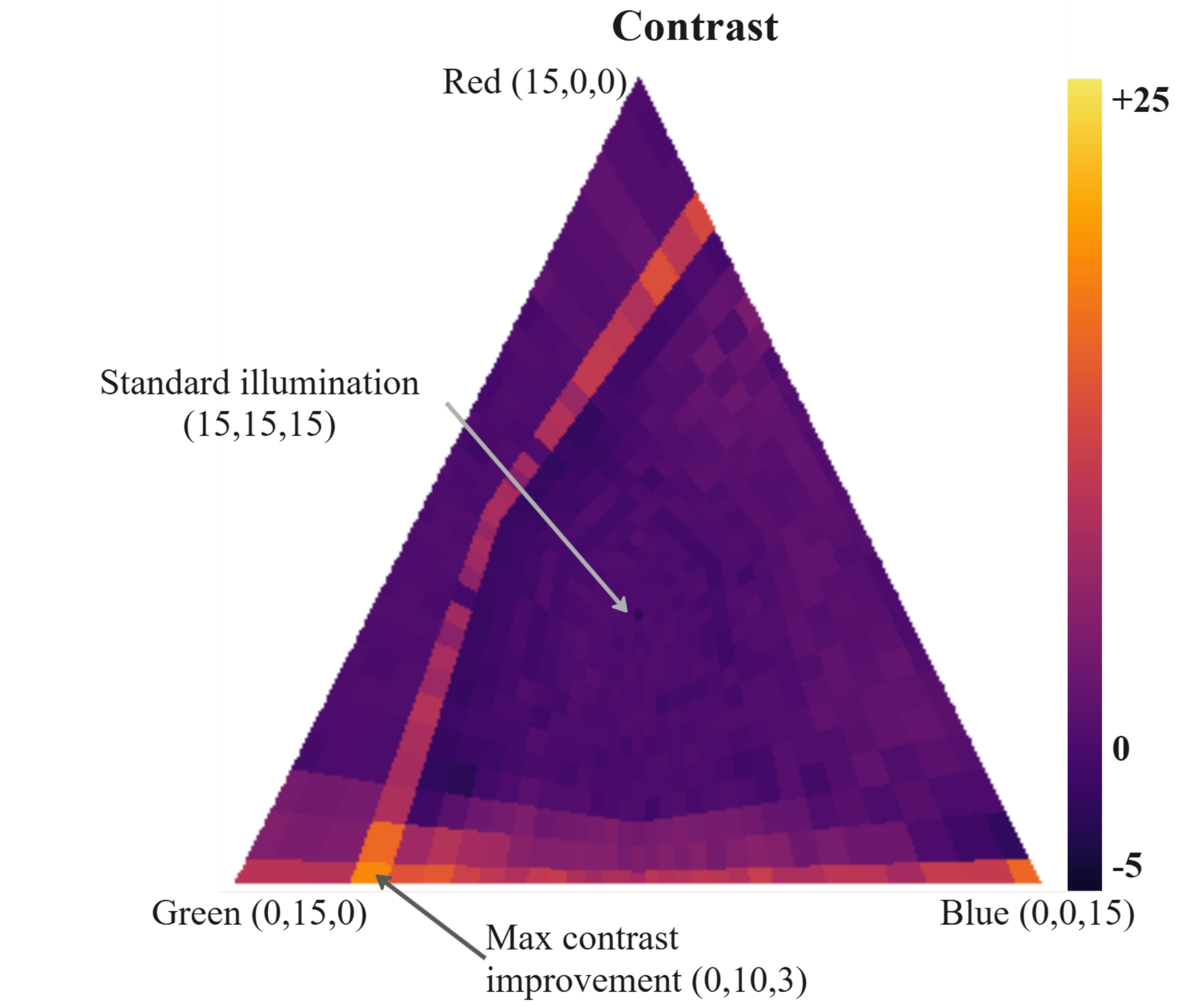}
    \end{subfigure}

  \caption{Heatmaps showing changes in contrast and sharpness of the image that is a result of image fusion of image taken with standard illumination and one more image taken with different illumination.   The greatest contrast increase was obtained when adding the image obtained with only green and blue LED lights on (0,10,3) and the greatest increase of sharpness was obtained setting intensities of RGB lights to (0,10,3).}
  \label{fig:triangle_heatmap}
\end{figure*}

\subsection{Proof-of-concept}
\begin{figure*}[t]
  \centering
  \includegraphics[width=0.85\linewidth]{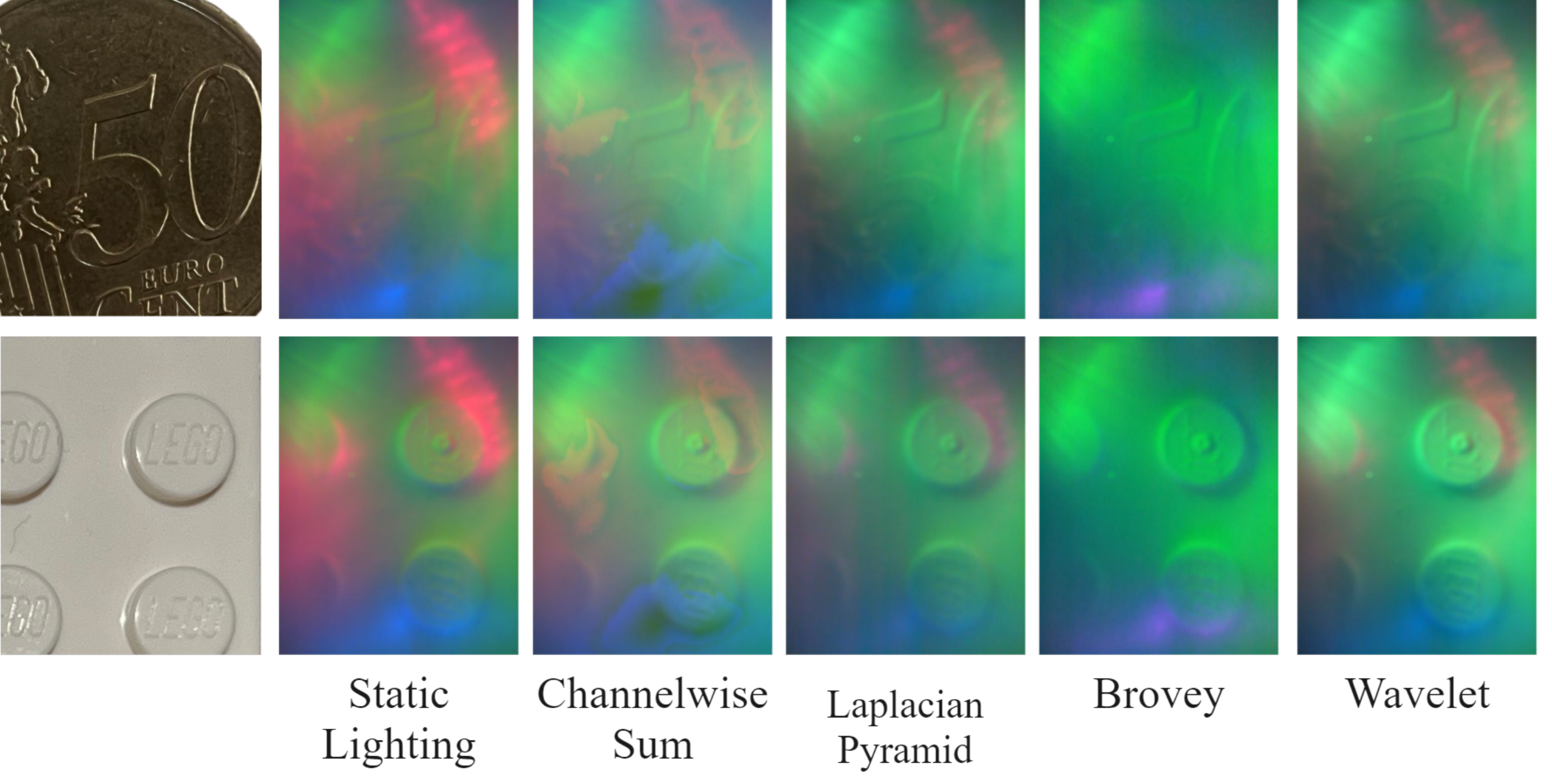}
  \caption{Measurements of a coin and Lego brick obtained using dynamic illumination and various image fusion methods}
  \label{figure:objectillumination}
\end{figure*}

In the first experiment, we ask the question: \textit{Can measurements acquired under standard illumination  be improved by combining them with images captured under different illumination?}
For this purpose, we selected a single object as shown in \fig{fig:coin_different_illumination}, and captured images under all possible sensor illumination settings, adjusting the intensities from 0 to 15. 
Each image was then combined with a reference image taken under standard DIGIT illumination (15,15,15) using DWT Image Fusion as an image fusion technique. 
This process resulted in a set of fused images, each representing the combination of two measurements (\fig{figure:objectillumination}). 
We calculated the contrast and sharpness for each fused image. 
The results indicate that fusing an image taken under standard illumination with images captured under different lighting conditions can improve these metrics, suggesting that dynamic illumination can be a valuable approach to enhancing image quality. 
On the heatmap in \fig{fig:triangle_heatmap} one can see how the lighting settings under which the second measurement was made affected the quality of the resulting image compared to the quality of the image obtained under standard static lighting. 
The greatest contrast increase was obtained when adding the image obtained with only green and blue LED lights on (0,10,3) and the greatest increase of sharpness was obtained setting intensities of RGB lights to (0,10,3). 
Thus, the measurements acquired under standard illumination  be improved by combining them with images captured under different illumination. 
This provides a basis for further exploration and indicates that combining images captured under different illumination conditions could be effective.

\subsection{Data Collection}
We then collected a larger dataset from all the eight object.
For each object, data collection comprised of two steps:
1) Background image collection: For each illumination setting represented by a tuple of intensities (r,g,b), we set the DIGIT illumination intensities to (r,g,b) and collected 100 images without any contact with the objects. We then calculated the average of these images to obtain the resulting background image.
2) Object image collection: We placed the object in contact with the sensor and, for each intensity tuple (r,g,b), set the DIGIT illumination intensity to (r,g,b). 
Overall, for each of the 23 illumination settings determined by tuples of intensities (r,g,b) of the LED lights, we obtained a background image and an image of each of the 9 objects (we treat the two sides of the coin as two different objects).

\subsection{Enhancing Image Quality}
Then, we ask: is it feasible to enhance image quality from the DIGIT sensor using dynamic lighting and image fusion techniques?
To assess this, we captured images of various objects under different illuminations. 
Initially, we employed the channel-wise summation method, alternating between illuminating the objects solely with red, green, and blue light intensities of (15,0,0), (0,15,0), and (0,0,15), respectively. 
Subsequently, we expanded our measurements to include additional illumination settings such as (15, 15, 0), (0, 15, 15), (15, 10, 5), and so forth, to leverage the Laplacian pyramid image fusion method. 
This method was then applied to sets of 2 and 3 images. 
To demonstrate the effectiveness of the Laplacian pyramid method, we combined images taken with intensity settings (15,15,0) and (0,0,15).

Upon analysis, we found that the Laplacian pyramid method consistently improved image quality for all objects in terms of background difference and contrast for most of the objects compared to images taken with standard DIGIT illumination settings. 
Meanwhile, the channel-wise sum method improved background difference and sharpness for all objects, as well as contrast for most objects.

Consequently, the utilization of dynamic lighting and image fusion techniques led to enhanced image quality obtained with the DIGIT vision-based tactile sensor.  
Thus, it is feasible  to enhance image quality from the DIGIT sensor using dynamic lighting and image fusion techniques

\subsection{Metrics and the most effective method}
\label{sec:metrics}
With the understanding that image quality enhancement is attainable through dynamic lighting and image fusion techniques, our subsequent investigation delves into two main inquiries. 
Firstly, we aim to ascertain whether metrics are correlated and is it is plausible to improve all metrics at once? 
Secondly, we seek to identify: what is the most effective method that can optimize all metrics simultaneously for all objects?
To address these questions, we conducted additional experiments involving different illumination settings and applied various image fusion techniques.

For each possible combination of 1 to 5 different illumination settings ${\theta_1, \ldots, \theta_i}$, where $\theta_j = (r_j, g_j, b_j)$, the corresponding set of images was selected. 
Subsequently, each fusion technique was applied to obtain the resulting image $I^* = f(I_{\theta_1}, \ldots, I_{\theta_i})$. 
The metrics were then calculated for the resulting image $I^*$.
Through comprehensive analysis of metric values across methods, illumination combinations, and objects, we concluded that the DWT-based method demonstrates the highest likelihood of optimizing all metrics simultaneously.

\subsection{Experimental Results}
Then, for each metric, we identified the combinations of illuminations and fusion methods that yielded the highest values. 
This process generated sets comprising pairs $(\text{illuminations}, \text{fusion method})$ for each metric. 
We then extracted their intersection, resulting in a set of pairs representing the optimal combinations of illuminations and fusion methods for each metric. 
Subsequently, for each object, we obtained a set of $(\text{illuminations}, \text{fusion method})$ pairs that demonstrated the highest metric values across all metrics. 
Finally, we curated pairs that consistently provided high metric values across all objects, resulting in the ultimate set of pairs $(\text{illuminations}, \text{fusion method})$.

\begin{figure*}[t]
  \centering
  \includegraphics[width=\linewidth]{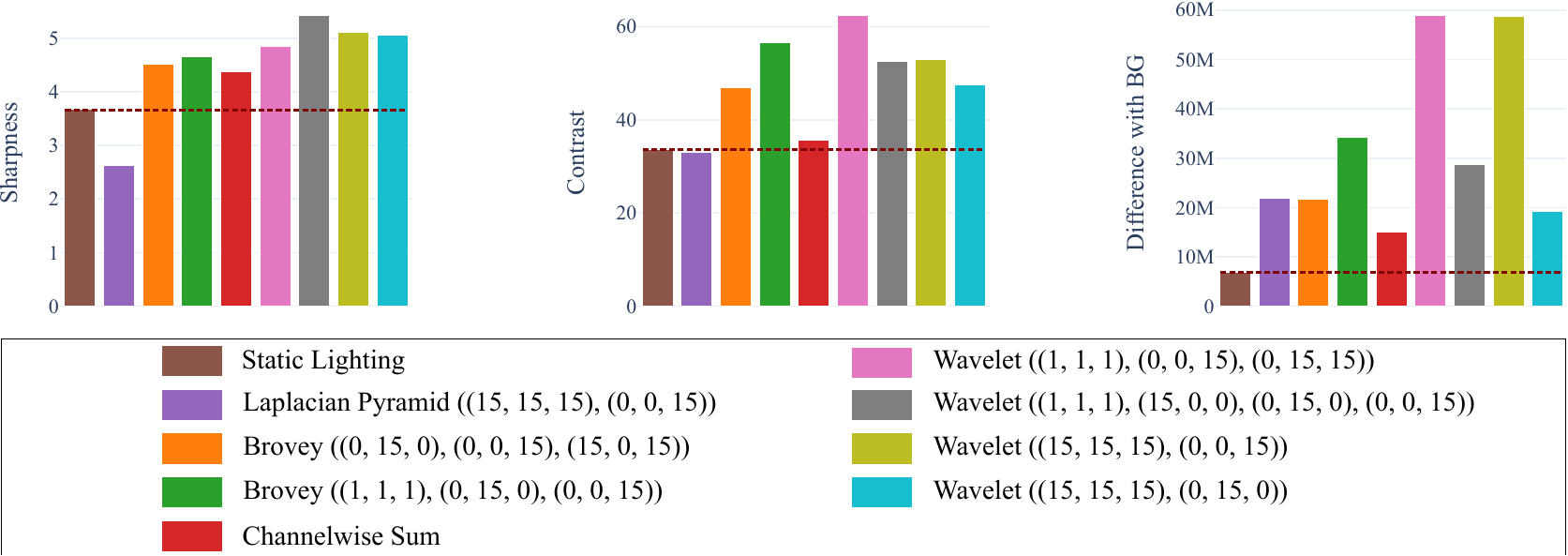}
  \caption{Average metrics over all of the objects. Laplacian pyramid method enhanced the difference with the background and contrast of the images. Channelwise sum method has improved both difference with the background and sharpness. Dynamic lighting with Wavelet and Brovey image fusion methods increased all of the metrics for all objects simultaneously. Overall, Wavelet methods provide the highest increase in image metrics.}
  \label{fig:metrics_objects}
\end{figure*}

Our findings reveal that applying dynamic illumination and image fusion techniques to images obtained from the sensor enhances image quality, improving contrast, sharpness, background difference, and human perception simultaneously. 
Particularly noteworthy is the discovery that the fusion method yielding the highest performance was the Wavelet Transform Image Fusion when applied to images acquired with (15,15,15) and (0,15,0) illumination intensity settings (see \fig{fig:metrics_objects}).

\subsection{Number of Images}

In our previous experiments, we fused from 2 to 4 images to generate a single output image of enhanced quality. 
This  raises the question: What is the optimal number of images to be fused? 
To address this, we conducted an additional experiment involving images of 130 different objects. 
Each object was captured under all possible illumination settings defined by $\theta = (r, g, b)$, where $r, g, b \in \{0, 1, 5, 15\}$.
For this experiment, we employed WDT Image Fusion, which demonstrated superior performance in our prior evaluations. 
We considered sequences of illumination settings $s^{(n)} = \{\theta_1, \theta_2, \dots, \theta_n\}$ of lengths up to 12. Sequence $s^{(1)\ *}=\{\theta_1^*\}$ of length 1 is optimal if
\[
\theta_1^*\ = \arg\max_{\theta_1} P(I_{\theta_1}),
\].
Each subsequent optimal sequence of length $m+1$ is formed by adding one more image to the previously determined optimal sequence of length $m$, denoted as $s^{(m)\ *} = \{\theta_1^*, \theta_2^*, \dots, \theta_m^*\}$. The extended sequence is:
\[
    s_{m+1}^* = \{\theta_1^*, \theta_2^*, \dots, \theta_m^*, \theta_{m+1}^*\},
\]
\[
    where \ \theta_{m+1}^*=\arg\max_{\theta_{m+1}} P\left(f(I_{\theta_1^*}, \dots, I_{\theta_m^*}, I_{\theta_{m+1}})\right),
\]
where $f(\cdot)$ denotes the image fusion.

Using this greedy strategy, we computed the contrast and sharpness of the optimal sequences of lengths from 1 to 12 for each of 130 objects. The optimal number $n^*$ of images to fuse for each object, was determined as a length of illumination settings sequence that maximizes metric $P$ .

\[
    n^* = \arg\max_{n\in\mathbb{N}} P(\mathop{\mathcal{F}}\limits_{i=1}^{n} (I_{\theta_i^*})),
\]
\[
   \mathop{\mathcal{F}}\limits_{i=1}^{n}(I_{\theta_i^*}) = f(I_{\theta_1^*},I_{\theta_2^*},..., I_{\theta_n^*}).
\]

\begin{figure*}[t]
  \centering
      \begin{subfigure}{0.49\linewidth}
        \centering
        \includegraphics[height=5.7cm]{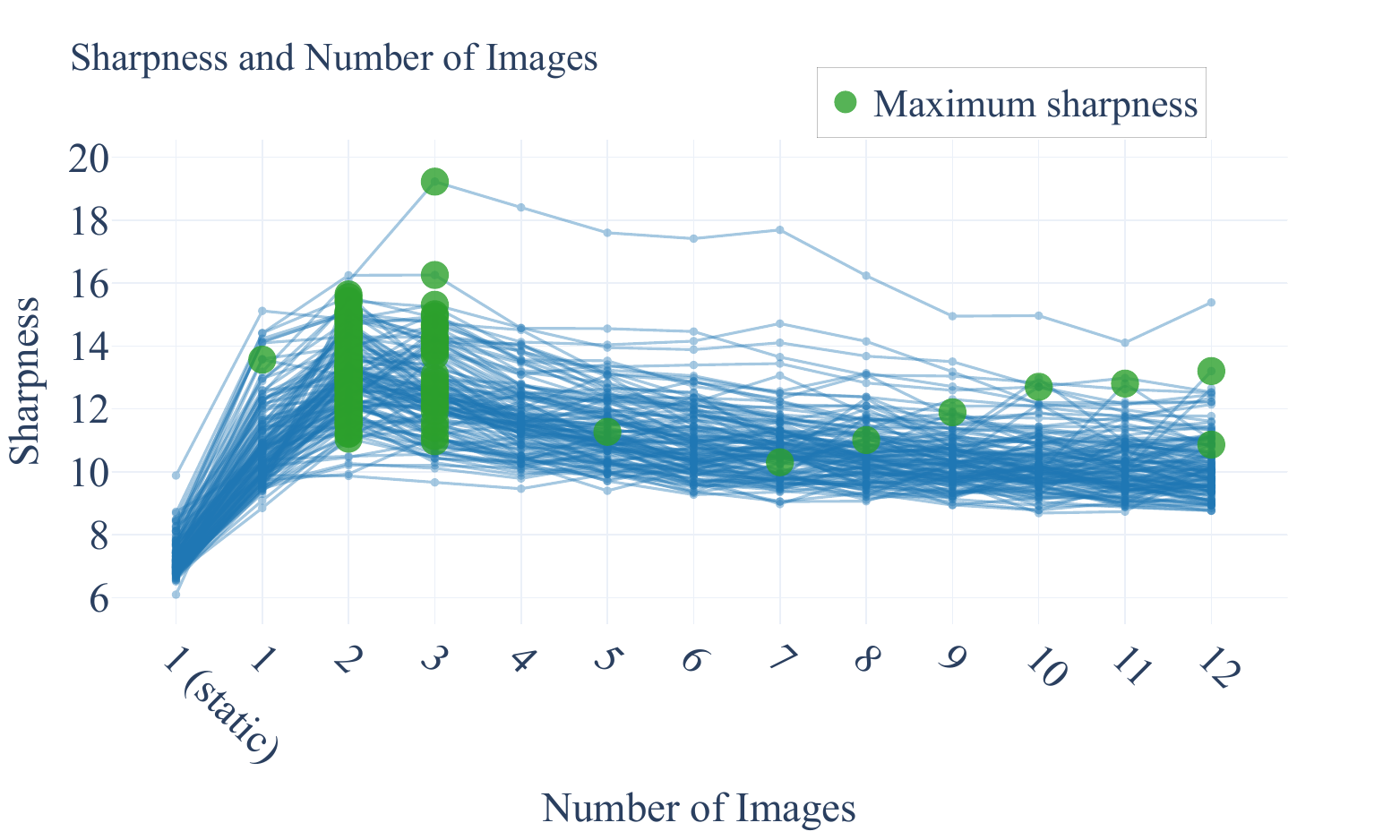}
    \end{subfigure}
    \hfill
    \begin{subfigure}{0.49\linewidth}
        \centering
        \includegraphics[height=5.7cm]{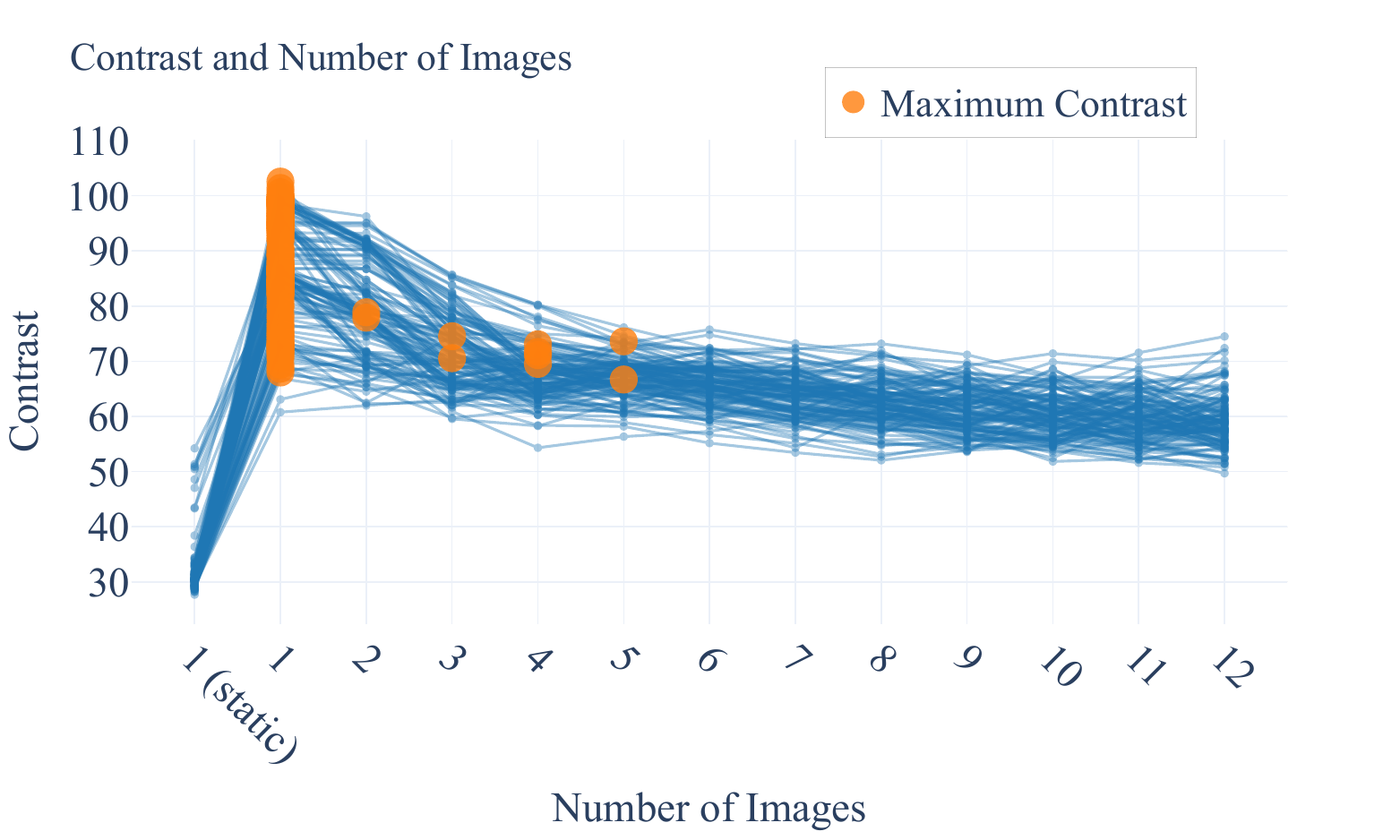}
    \end{subfigure}
  \caption{Charts showing maximum sharpness and contrast achieved for illumination settings sequences of different lengths for different objects, each line corresponds to one object. Colored circles denote points of maximum sharpness and contrast over all optimal sequences of length from 1 to 12 for each object. For most of the objects 2-3 images lead to maximum sharpness and just 1 image would be enough to achieve maximum contrast if the (unknown) optimal illumination is applied.}
  \label{fig:n_images}
\end{figure*}

\fig{fig:n_images} shows the effect of sequence length on the sharpness and contrast of the fused image.
We observe that, for the majority of objects, the optimal number of images for maximizing sharpness ranges between 2 and 4. 
In contrast, for maximizing image contrast, a single image often yields the best result. 
Adding more images beyond these points tends to degrade the quality of the fused image.

\subsection{Time and image quality}

\begin{figure}[t]
  \centering
\includegraphics[width=\linewidth]  {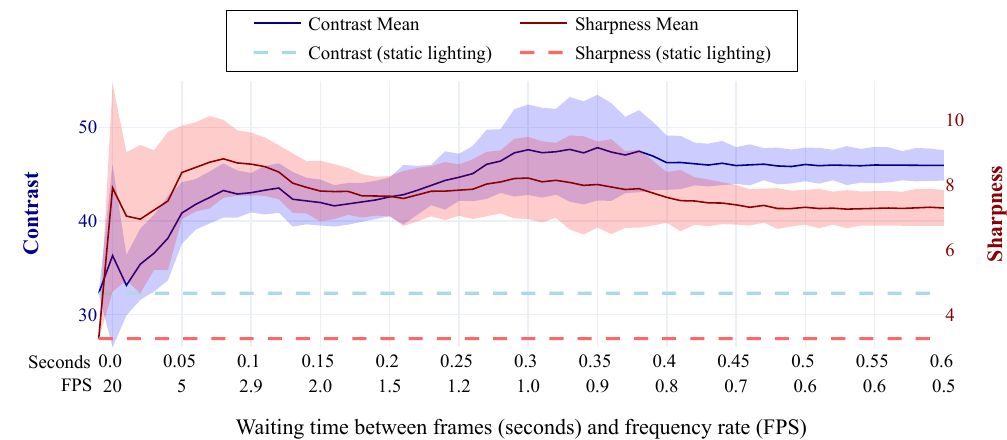}
  \caption{Mean and 95\% confidence of the contrast and sharpness against the time between frames. Dynamic lighting enhances the sharpness of the captured images regardless of the time between frames. However, when the interval is less than \SI{0.1}{\second}, the sharpness becomes unstable and shows high variance. Contrast increases as the time between frames lengthens. After approximately \SI{0.3}{\second} between frames, the contrast remains largely unchanged}
  \label{fig:ci_contrast_and_sharpness}
\end{figure}

Finally, we study the question: how much time is required to effectively apply dynamic lighting? 
To use dynamic lighting and get one resulting image, it is necessary to take several measurements one by one and change the lighting settings between them. 
The quality of the resulting measurements depends on the time between these frames. 
We conducted an experiment for which we used one object, used dynamic lighting and Wavelet image fusion algorithm for three lighting settings. 
We tested how changing the waiting time between frames from 0 to 0.6 seconds would affect the quality of the resulting image - its sharpness and contrast. 
For this purpose, we obtained 100 measurements for each waiting time.
As shown in \fig{fig:ci_contrast_and_sharpness}, for the low waiting time of \mbox{0-0.1} seconds, the metrics values obtained for 100 images have a high variance and thus the image quality is not stable. 
In addition, the images do not seem to be optimal for human eye perception. 
As the waiting time between frames increases, the variance decreases, sharpness and contrast become consistently higher than for images obtained with static illumination, and further the values reach a plateau.  
Thus, it turned out to be most effective to use dynamic lighting settings with \SI{0.29}{\second} between frames, which would result in a frame rate of \SI{1.1}{FPS}, when using 3 illumination settings.

%% file: 5_conclusion.tex
Traditional vision-based tactile sensors make use of static illumination patterns that are optimized at design time.
In this study, we instead propose to enhance images captured by vision-based tactile sensors using dynamic illumination and image fusion techniques. 
This progress paves the way for improved measurements across all vision-based tactile sensors where dynamic lighting is applicable, thereby boosting the overall accuracy and performance of robotic tasks that depend on these sensors. 
Experimental results demonstrated that dynamic lighting significantly improves measurement quality like contrast, sharpness, and background differentiation.
Among the several image fusion methods evaluated, we identified Discrete Wavelet Transform Image Fusion as the most effective technique for combining measurements from haptic sensors under different lighting conditions. 
%
Future work will focus on evaluating dynamic illumination on more complex sensors, such as the Digit360~\cite{Lambeta2024Digitizing} (which possess 8 fully controllable RGB LEDs), and extending the problem formulation to be object dependent.

%% file: 99_acknowledgments.tex
This work was partly supported by the German Research Foundation (DFG, Deutsche Forschungsgemeinschaft) as part of Germany’s Excellence Strategy – EXC 2050/1 – Project ID 390696704 – Cluster of Excellence “Centre for Tactile Internet with Human-in-the-Loop” (CeTI) of Technische Universität Dresden, and by Bundesministerium für Bildung und Forschung (BMBF) and German Academic Exchange Service (DAAD) in project 57616814 (\href{https://secai.org/}{SECAI}, \href{https://secai.org/}{School of Embedded and Composite AI}).